\pdfoutput=1
\documentclass[11pt]{article}

\usepackage[preprint]{acl}
\usepackage{times}
\usepackage{latexsym}
\usepackage{enumitem}
\usepackage{amsmath}
\usepackage{adjustbox}
\usepackage{booktabs}
\usepackage{multirow}
\usepackage{hyperref}
\usepackage[T5]{fontenc}
\usepackage[vietnamese,english]{babel}

\usepackage[utf8]{inputenc}

\usepackage{microtype}

\usepackage{inconsolata}

\usepackage{graphicx}

\title{Multilingual Test-Time Scaling via Initial Thought Transfer}

\author{First Author \\
  Affiliation / Address line 1 \\
  Affiliation / Address line 2 \\
  Affiliation / Address line 3 \\
  \texttt{email@domain} \\\And
  Second Author \\
  Affiliation / Address line 1 \\
  Affiliation / Address line 2 \\
  Affiliation / Address line 3 \\
  \texttt{email@domain} \\}

\author{Prasoon Bajpai \\
  IIT Delhi \\
  \texttt{prasoonbajpai786@gmail.com} \\\And
  Tanmoy Chakraborty \\
  IIT Delhi \\
  \texttt{tanchak@iitd.ac.in} \\}

\begin{document}

\maketitle

\begin{abstract}
Test-time scaling has emerged as a widely adopted inference-time strategy for boosting reasoning performance. However, its effectiveness has been studied almost exclusively in English, leaving its behavior in other languages largely unexplored. We present the first systematic study of test-time scaling in multilingual settings, evaluating \texttt{DeepSeek-R1-Distill-LLama-8B} and \texttt{DeepSeek-R1-Distill-Qwen-7B} across both high- and low-resource Latin-script languages. Our findings reveal that the relative gains from test-time scaling vary significantly across languages. Additionally, models frequently switch to English mid-reasoning, even when operating under strictly monolingual prompts. We further show that low-resource languages not only produce initial reasoning thoughts that differ significantly from English but also have lower internal consistency across generations in their early reasoning. Building on our findings, we introduce MITT (Multilingual Initial Thought Transfer), an unsupervised and lightweight reasoning prefix-tuning approach that transfers high-resource reasoning prefixes to enhance test-time scaling across all languages, addressing inconsistencies in multilingual reasoning performance. MITT significantly boosts \texttt{DeepSeek-R1-Distill-Qwen-7B}’s reasoning performance, especially for underrepresented languages. \footnote {The dataset and code can be found in the github repository \href{https://github.com/Prasoon1207/multilingual-test-time-scaling}{Link}}
\end{abstract}

\begin{figure}[h]
    \centering
    \includegraphics[width=\linewidth]{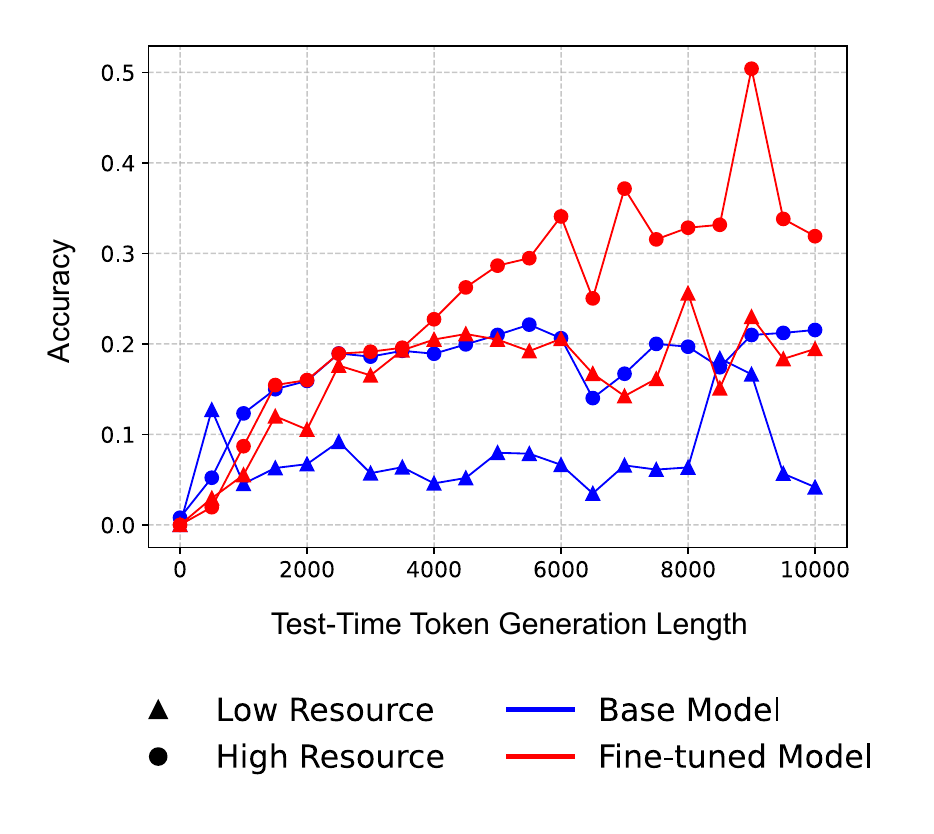}
    \caption{Effect of Multilingual Initial Thought Transfer (\textbf{MITT}) on \texttt{DeepSeek-R1-Distill-Qwen-7B}. We observe that fine-tuning on initial reasoning steps in English (<32 tokens) offers an unsupervised and data efficient way to not only improve accuracy but also enhance progressive gains from scaling test-time compute in low as well as high resource language settings.}
    \label{fig:resource_qwen_e3}
\end{figure}
\begin{figure*}
    \centering
    \includegraphics[width=0.88\linewidth]{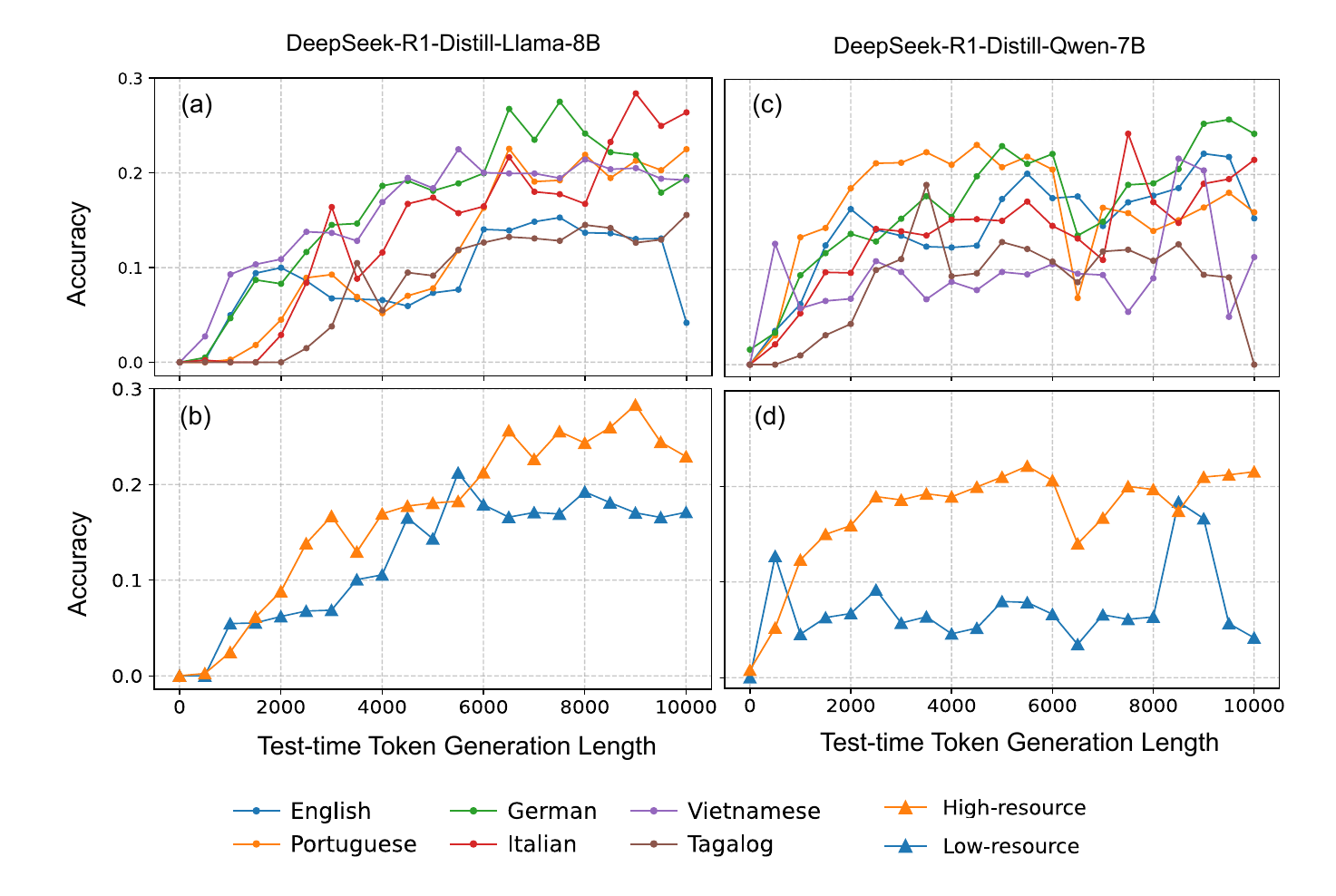}
    \vspace{-0.5cm}
    \caption{Test-time scaling trends for (a) + (b) \texttt{DeepSeek-R1-Distill-LLama-8B}  and (c) + (d) \texttt{DeepSeek-R1-Distill-Qwen-7B}: . (a) and (c) display overall trends in test-time scaling across all languages, while (b) and (d) present average gains separately for low-resource and high-resource language groups. The results reveal a consistent pattern: both models demonstrate stronger test-time scaling in high-resource languages compared to low-resource ones. \texttt{DeepSeek-R1-Distill-Qwen-7B} shows insignificant test-time scaling for low-resource languages.}
    \label{fig:overall_resource}
\end{figure*}
\section{Introduction}

Large Language Models (LLMs) acquire broad capabilities through scaling training-time compute, where they learn from increasing volumes of data and parameters \cite{bubeck2023sparksartificialgeneralintelligence, jones2025largelanguagemodelspass}. However, this pretraining paradigm faces diminishing returns due to its high resource demands and the finite availability of high-quality human data \cite{cottier2025risingcoststrainingfrontier, villalobos2024rundatalimitsllm}. Recently, focus has shifted toward test-time scaling \cite{deepseekai2025deepseekr1incentivizingreasoningcapability, muennighoff2025s1simpletesttimescaling}, a phenomenon where a model’s reasoning performance improves as the length of its generated output increases during inference, revealing latent reasoning capabilities without the need for additional training. Test-time scaling is model-agnostic, inference-time-only, and requires no gradient updates, making it attractive for scaling performance without retraining. Understanding its limitations across languages is essential for ensuring multilingual generalization, especially in real-world settings where consistent reasoning across languages is critical. It remains unclear whether this phenomenon generalizes consistently across languages in strictly monolingual settings. In this work, we ask: \textit{Does test-time scaling manifest equally across languages in strictly monolingual settings? And how do factors influence disparities in multilingual scaling behavior, if any?}

To answer this question, we conduct a systematic investigation using two state-of-the-art models: \texttt{DeepSeek-R1-Distill-LLama-8B} and \texttt{DeepSeek-R1-Distill-Qwen-7B} \cite{deepseekai2025deepseekr1incentivizingreasoningcapability}. We evaluate their long-horizon reasoning behavior in strictly monolingual contexts across a diverse set of high- and low-resource Latin-script languages.\par
We address the following research questions:
\begin{itemize}[noitemsep, topsep=0pt, left=0pt]
    \item \textbf{RQ1}: How does test-time scaling manifest across different languages, particularly for low-resource languages?
    \item \textbf{RQ2}: How do reasoning patterns, especially the initial ``thoughts” differ across languages?
    \item \textbf{RQ3}: Are low-resource languages internally consistent in their initial reasoning patterns, or do they exhibit greater variability across generations? How similar are these patterns compared to other languages?
    \item \textbf{RQ4}: Can certain targeted adaptation strategies improve reasoning gains through test-time scaling in under-performing models and languages?
\end{itemize}

To answer these research questions, we conduct a comprehensive empirical analysis of multilingual test-time scaling behavior. (\textbf{Contribution 1}) As the foundational step, we curate low-resource Latin-script translations of the AIME 2025 dataset \cite{AIME2025} in Vietnamese and Tagalog (Filipino), extending the multilingual benchmark to better reflect the challenges of underrepresented languages in structured reasoning tasks. (\textbf{Finding 1}) We find that while \texttt{DeepSeek-R1-Distill-LLama-8B} demonstrates consistent gains in accuracy with increased generation length across both high- and low-resource languages, \texttt{DeepSeek-R1-Distill-Qwen-7B} exhibits significantly more erratic and inconsistent scaling behavior, particularly for low-resource settings. (\textbf{Finding 2}) Our analysis also reveals a surprising degree of language flipping during long reasoning outputs: even in monolingual contexts, models often begin reasoning in the target language but switch mid-stream to English, with this effect being especially prominent in \texttt{DeepSeek-R1-Distill-Llama-8B} and in low-resource language outputs. (\textbf{Finding 3}) To better understand these disparities in test-time scaling behavior, we analyze the similarity of initial reasoning patterns across languages and find that low-resource languages not only diverge more sharply from English but also exhibit lower intra-language consistency in their initial reasoning patterns, suggesting less stable initial reasoning structures. (\textbf{Contribution 2}) Finally, we introduce \textbf{Multilingual Initial Thought Transfer (MITT)}, a data-efficient paradigm for improving reasoning through test-time scaling in multilingual models. Instead of tuning directly on target-language data, MITT leverages unsupervised prefix-tuning using initial reasoning chains generated in a high-resource language. This method transfers structured inductive priors across languages, enabling \texttt{DeepSeek-R1-Distill-Qwen-7B} to achieve significantly more stable and effective test-time scaling, particularly in low-resource settings (Figure \ref{fig:resource_qwen_e3}). MITT offers a lightweight and model-agnostic intervention for enhancing multilingual generalization without the need for supervised cross-lingual data. Together, this study exposes key structural gaps in current multilingual models and offers both diagnostic tools and practical interventions for improving multilingual reasoning fidelity.

\begin{figure*}[ht]
    \centering
    \includegraphics[width=0.8\linewidth]{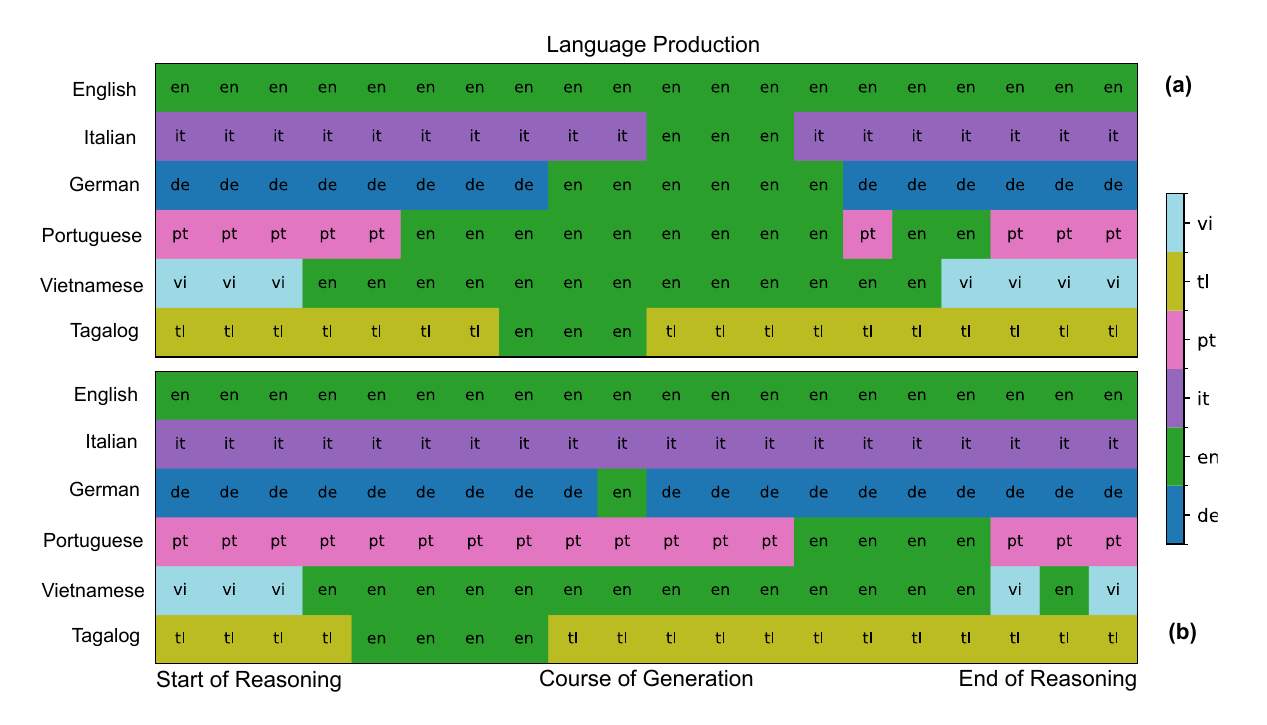}
    \caption{Visualization of the sentence-level language trajectory across the generated reasoning stream. Each row represents the dominant language detected at each generation segment, with panels \textbf{(a)} and \textbf{(b)} corresponding to \texttt{DeepSeek-R1-Distill-LLama-8B} and \texttt{DeepSeek-R1-Distill-Qwen-7B}, respectively. While the models are prompted monolingually in the target language, we observe the intrusion of English into the generation process.}
    \label{fig:lang_id_2}
\end{figure*}
\begin{figure}[h]
    \centering
    \includegraphics[width=\linewidth]{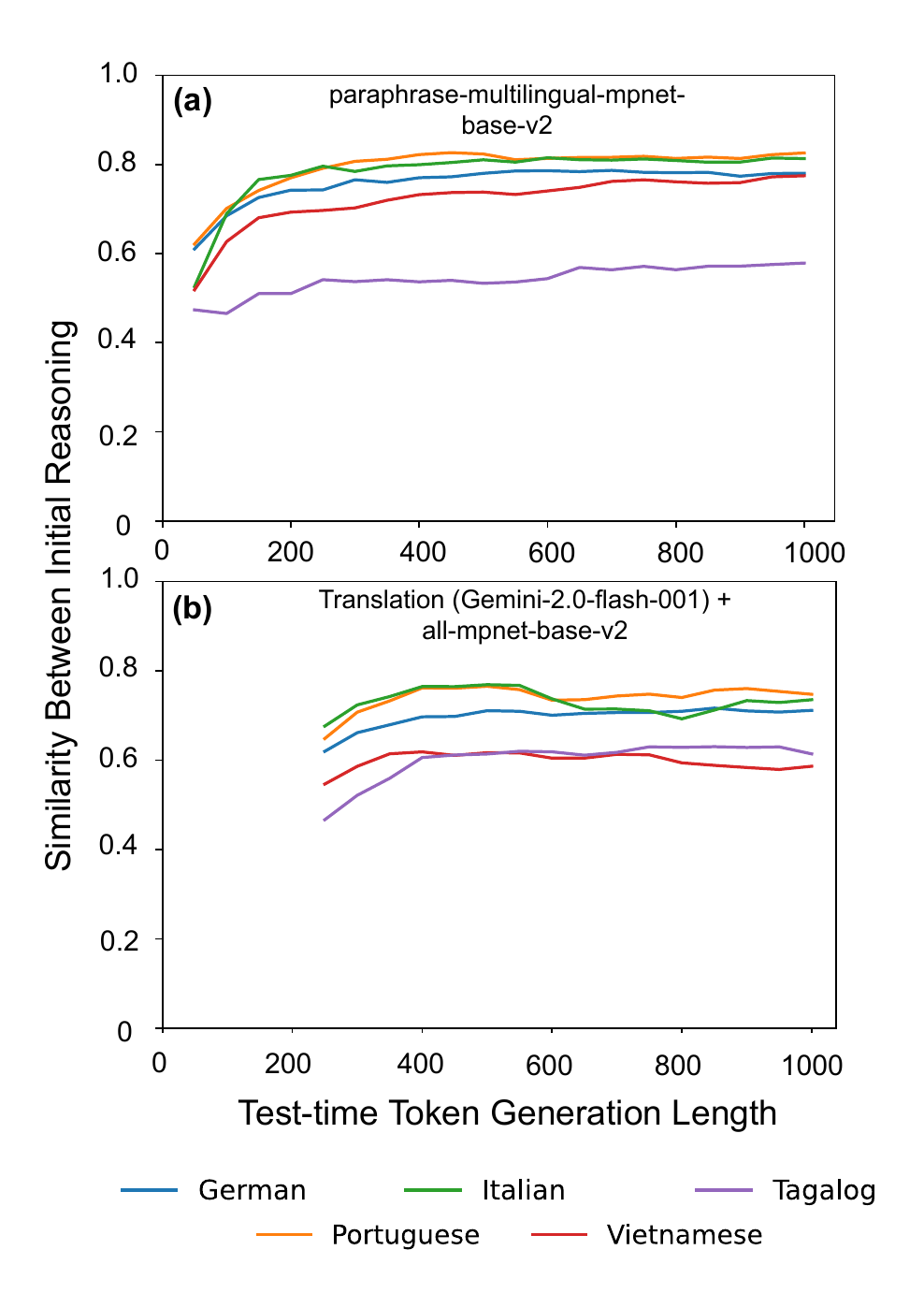}
    \vspace{-0.3cm}
    \caption{Comparison of the similarity of initial reasoning segments (upto 1000 tokens) between English and other target languages. To enhance interpretability, we apply a rolling average with a window size of 5 in (b).}
    \label{fig:similarity_initial_reasoning}
\end{figure}
\begin{figure*}[ht]
    \centering
    \includegraphics[width=0.95\linewidth]{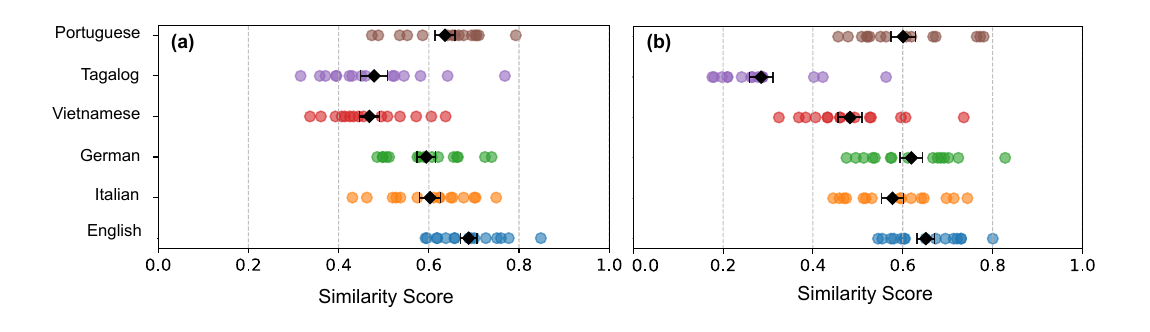}
    \caption{Distribution of similarity scores across all questions, for initial reasoning segments (first 32 tokens) sampled from 100 generations per question across six languages. All reasoning chains are first translated to English using Gemini and then embedded using an English monolingual encoder to compute pairwise intra-language similarity.  \textbf{(a)} Results for \texttt{DeepSeek-R1-Distill-LLama-8B}, and  \textbf{(b)} results for \texttt{DeepSeek-R1-Distill-Qwen-7B}.}
    \label{fig:intra_prefix_similarity}
\end{figure*}
\section{Related Work}
\textbf{Test-time Scaling.} The paradigm of test-time scaling \cite{ji2025testtimecomputesystem1thinking, zhang2025surveytesttimescalinglarge} has gained prominence as an alternative to traditional pretraining scaling, aiming to enhance LLM performance during inference without modifying model parameters. The emergence of reasoning-specialized LLMs such as \texttt{DeepSeek-R1} \cite{deepseekai2025deepseekr1incentivizingreasoningcapability} and \texttt{OpenAI-o1} \cite{openai2024openaio1card} has marked a significant shift in how test-time reasoning is approached. These models extend the chain-of-thought (CoT) paradigm \cite{wei2023chainofthoughtpromptingelicitsreasoning} by enabling long structured sequences of intermediate logical steps, generated at inference time. These models provide opportunities for self-correction \cite{kumar2024traininglanguagemodelsselfcorrect, kamoi-etal-2024-llms}, where models revise flawed reasoning to arrive at more logical conclusions. This behavior exemplifies a form of test-time scaling, where richer inference traces yield stronger final outputs without any change to model parameters. Their success has catalyzed a new class of reasoning-focused strategies that aim to mimic \cite{xie2025logicrlunleashingllmreasoning, li2025draftsanswersunlockingllm, saunshi2025reasoninglatentthoughtspower} or distill long reasoning trajectories from larger teacher models \cite{muennighoff2025s1simpletesttimescaling, li2025llmseasilylearnreason}. Most prior work on test-time scaling has been heavily centered on English \cite{yang2025thinkingoptimalscalingtesttimecompute, teng2025atomthoughtsmarkovllm, lifshitz2025multiagent, snell2024scalingllmtesttimecompute, liu20251bllmsurpass405b}, with limited investigation into how such methods generalize across languages. This English-centric focus leaves open questions about the applicability of long-form reasoning and compute scaling in multilingual contexts. In our work, we extend this line of inquiry to a diverse set of Latin-script languages by evaluating test-time scaling in strictly monolingual settings, isolating each language to assess reasoning performance without any cross-lingual interference.\par

\textbf{Multilingual Reasoning}. Reasoning in LLMs, particularly via CoT prompting, has been primarily explored in English \cite{plaat2024reasoninglargelanguagemodels, li202512surveyreasoning}. Extensions to multilingual contexts remain limited \cite{shi2022languagemodelsmultilingualchainofthought, ghosh2025multilingualmindsurvey}. Multilingual CoT prompting has been studied in cross-lingual transfer settings \cite{qin-etal-2023-cross}, showing limited transferability of CoT traces from English to other languages, particularly low-resource ones. \par
Most reasoning benchmarks are English-centric (e.g., GSM8K \cite{cobbe2021trainingverifierssolvemath}, StrategyQA \cite{geva2021didaristotleuselaptop}), limiting insights into multilingual capabilities. To address this, new multilingual benchmarks have emerged. For instance, XCOPA \cite{ponti-etal-2020-xcopa} evaluates causal commonsense reasoning in 11 typologically diverse languages, while XWinograd\cite{muennighoff2023crosslingualgeneralizationmultitaskfinetuning} extends Winograd \cite{kocijan2020reviewwinogradschemachallenge} schema tasks. Meanwhile, MGSM \cite{shi2022languagemodelsmultilingualchainofthought} offers multilingual extensions of GSM8K, designed to test arithmetic and symbolic reasoning under language constraints. In contrast, we use a multilingual version of the  AIME question bank \cite{AIME2025}, an inherently challenging benchmark requiring multi-step mathematical reasoning. Its long-form and difficult nature makes it well-suited for studying test-time scaling.

% However, coverage of low-resource and morphologically complex languages remains sparse.
\par
Despite these efforts, no prior work has systematically assessed test-time scaling behavior of multilingual language models in strictly monolingual settings, where both the input and reasoning context are entirely in the target language without reliance on translation, multilingual prompts, or cross-lingual support. Our work fills this gap by conducting a controlled comparison of scaling effects across languages using monolingual contexts; analyzing intra- and inter-language CoT similarity  and introducing an unsupervised prefix-tuning paradigm that uses any high-resource language's reasoning priors to improve scaling behavior in low-resource languages without altering the test-time monolingual constraint.
\section{Experiment Setup}
Below, we outline the experimental design and implementation details.
\subsection{Models and Dataset}  
We conduct all experiments using two instruction-tuned, multilingual models from the DeepSeek-R1 family: \texttt{DeepSeek-R1-Distill-LLama-8B} and \texttt{DeepSeek-R1-Distill-Qwen-7B} \cite{deepseekai2025deepseekr1incentivizingreasoningcapability}. Both models are distilled and fine-tuned for general-purpose reasoning tasks but differ significantly along the axis of the base model's pretraining corpora composition \cite{grattafiori2024llama3herdmodels, bai2023qwentechnicalreport}. Models from the \texttt{LLama} family are pretrained on a broad and diverse multilingual corpus. In contrast, models from the \texttt{Qwen} family incorporate a more code-heavy and English-skewed pretraining mix with higher exposure to web-based and structured reasoning content. 

% These differences allow us to isolate the effects of pretraining linguistic bias on multilingual test-time scaling behavior.

We use the multilingual AIME 2025 dataset \cite{AIME2025}, which consists of two contest subsets, each containing 15 numerical mathematics questions. To ensure broad question-type coverage, we include both subsets in our experiments. Each question is originally available in four Latin-script languages. To expand the dataset’s multilingual scope, we augment it with translations in two additional low-resource Latin-script languages: Vietnamese and Tagalog (Filipino). We create these translations from English by using \texttt{GPT-4o} \cite{openai2024gpt4ocard} as the zero-shot translator. In this work, we classify languages as high-resource or low-resource based on their prevalence in the pretraining corpora of LLMs and their general availability in multilingual NLP benchmarks. According to this classification, we consider English, Italian, German, and Portuguese as high-resource, and Vietnamese and Tagalog as low-resource.

% Specifically, we treat a language as high-resource if it is widely represented in training datasets (e.g., Common Crawl, Wikipedia, code or instructional corpora) and appears frequently in cross-lingual benchmarks. Conversely, low-resource languages are those with limited representation in these corpora and minimal coverage in mathematical or reasoning-specific benchmarks. 

\subsection{Test-time Scaling Setup}

Let $Q^l$ denote a question in language $l$, $\mathcal{M}$ the model under study, $D^l$ a single demonstration in language $l$, and $S$ the instruction prompt. We construct the input prompt as a concatenation of these components: $[S, D^l, Q^l]$. The model $\mathcal{M}$ is then prompted to generate a reasoning trace:
\begin{equation*}
    R \sim \mathcal{M}([S, D^l, Q^l]) = (r_0, r_1, \dots, r_{T-1})
\end{equation*}
where $r_i$ denotes the $i^{\text{th}}$ token generated by the model and $T = 10000$ is the maximum number of generated tokens. To evaluate test-time scaling behavior, we periodically extract answers from the evolving reasoning trace. Specifically, after every 32 generated tokens, we append an answer extraction prompt $A$ and elicit a numerical answer $a_k$ from the model:
\begin{equation*}
a_k = \mathcal{M}([r_0, r_1, \dots, r_{32k-1}, A])\hspace{1mm}\text{for } k \in \mathbf{N} \cup \{0\}
\end{equation*}
\noindent
\text{where} $32k < T$. Each $a_k$ represents the model’s best answer based on the reasoning developed up to token position $32k$. This procedure allows us to trace the progression of reasoning quality over the course of generation and measure how answer correctness evolves as a function of reasoning length in language $l$.
\noindent
The answer extraction prompt is placed outside the generated reasoning text stream and hence, does not interfere with the current chain of thought. We use the Exact-Match metric to quantify the accuracy based on current reasoning.\par
Moreover, following recent works \cite{muennighoff2025s1simpletesttimescaling} on enforcing long reasoning chain generation, we insert a language-specific wait prompt $W^l$ (e.g., `\textit{Let me re-think my reasoning from scratch}' in English), whenever $\mathcal{M}$ tries to produce an answer on its own. The wait prompts forms a part of the reasoning text stream but it is not counted as a part of the newly generated tokens. More details on the exact prompt and inference method can be found in Appendix \ref{sec:prompt_details}. Moreover, we provide the set of all language-specific wait prompts, system prompts and answer extraction prompts in the Appendix, Tables \ref{tab:wait_prompt}, \ref{tab:system_prompt} and \ref{tab:answer_prompt}. We
show the results in Figure \ref{fig:overall_resource} for different models across all languages under study.

\subsection{Language Fidelity in Reasoning Chains}

To assess the language fidelity within long reasoning chains, we segment the generated token sequence into fixed-length windows and assign a language label to each window. Let $R = \{r_0, r_1, \dots, r_{T-1}\}$ denote the full reasoning trace comprising $T$ tokens. We partition $R$ into non-overlapping windows of length 32 tokens:

\begin{equation*}
R = \bigcup_{k=0}^{\lfloor T / 32 \rfloor - 1} W_k, \quad
\end{equation*}
where, $\quad W_k = \{r_{32k}, \dots, r_{32(k+1) - 1}\}$. For each window $W_k$, we apply the \texttt{langid} \cite{lui-baldwin-2011-cross} classifier to obtain a predicted language label: $l_k = \texttt{langid}(W_k)$.

This yields a sequence of predicted languages $\{l_0, l_1, \dots, l_{m-1}\}$, where $m = \lfloor T / 32 \rfloor$.
To create a coarse-grained summary of language usage over the full reasoning trace, we divide this sequence into 20 equally sized segments $\{S_i\}_{i=1}^{20}$ and compute the majority language in each segment. We define the down-sampled language trace as:
\begin{equation*}
L^{\text{maj}} = \{\text{mode}(S_0), \text{mode}(S_1), \dots, \text{mode}(S_{19})\}
\end{equation*}
This representation enables visualization and analysis of language drift patterns along the reasoning trajectory. We plot these representations for each target language in Figure \ref{fig:lang_id_2}. We provide an alternative, more fine-grained analysis in Appendix \ref{sec:language_fidelity}.

\subsection{Similarity Analysis of Initial Reasoning Segments}

For each question $Q^l$, we extract the first 1000 tokens of the generated reasoning trace, denoted as $[r_0, r_1, \dots, r_{999}]^l$. This sequence is then divided into incremental segments of length 32 tokens. Specifically, for each integer $k \geq 0$ such that $32(k+1) \leq 1000$, we define the $k$-th reasoning segment as: $R_k^l = [r_{0}, \dots, r_{32(k+1)-1}]^l$. We compare these reasoning segments across languages by evaluating the similarity between each $R_k^l$ in a non-English language and its English counterpart. This comparison is conducted for all valid values of $k$, using two different similarity assessment methods:
\begin{enumerate}[noitemsep, topsep=0pt, left=0pt]
    \item \textbf{Multilingual Embedding Comparison:} Each segment $R_k^l$ is embedded using \texttt{paraphrase-multilingual-mpnet-base-v2} \cite{reimers-2020-multilingual-sentence-bert} to compute cosine similarity with the corresponding English segment $R_k^{en}$. 
    
    \item \textbf{Translation-based Embedding Comparison:} Each $R_k^l$ is first translated to English using \texttt{Gemini-2.0-flash-001}, and then embeddings are obtained using \texttt{all-mpnet-base-v2} \cite{reimers-2019-sentence-bert} for similarity computation.
\end{enumerate}

Using both multilingual and translation-based embeddings helps mitigate biases specific to any single embedding model, providing a more robust measure of cross-lingual initial reasoning similarity. The resulting similarity trends are presented in Figure~\ref{fig:similarity_initial_reasoning}. To enhance interpretability, we apply a rolling average with a window size of 5 in Figure~\ref{fig:similarity_initial_reasoning}(b).

\subsection{Intra- and Inter-Language Consistency of Reasoning Prefixes}

For each question $Q^l$ in language $l$, we sample 100 independent reasoning traces, each limited to the first 32 generated tokens. The goal is to assess the consistency of the initial reasoning (referred to as the \textit{reasoning prefix}) in relation to the resource availability of the language. To compute similarity among reasoning prefixes within the same language, we avoid potential multilingual embedding biases by using the two-step approach: each prefix is first translated into English using \texttt{Gemini-2.0-flash-001}, followed by embedding with \texttt{all-mpnet-base-v2} to compute cosine similarity. Figure~\ref{fig:intra_prefix_similarity} presents the intra-language consistency results. For each language, we compute pairwise similarity among the 100 reasoning prefixes for each question, then average these to obtain a question-level consistency score. Finally, we report the mean of these scores across all questions to obtain a language-level measure of prefix consistency. We also quantify cross-lingual consistency of reasoning prefixes. For each question, we compute the pairwise similarity between reasoning prefixes from different languages, average these values to obtain a language-pair score, and then aggregate across all questions. The resulting inter-language similarity matrix is shown in Figure~\ref{fig:inter_prefix_similarity} (Appendix).

\subsection{MITT: Intervention via Unsupervised Prefix-Tuning}
Motivated from a study \cite{https://doi.org/10.13140/rg.2.2.33772.07043} that shows the impact of initial reasoning patterns on long-horizon reasoning performance, we challenge a similar hypothesis in a cross-lingual scenario. To address the inconsistent test-time scaling behavior observed in \texttt{DeepSeek-R1-Distill-Qwen-7B}, we propose Multilingual Initial Thought Transfer (MITT), a cross-lingual, unsupervised prefix-based fine-tuning approach. We experiment with two training strategies: (1) collecting all English reasoning prefixes across questions and fine-tuning for three epochs using the standard causal language modeling objective, and (2) aggregating prefixes from all high-resource languages (English, Italian, German, and Portuguese) and training for one epoch. Specifically, we extract all 100 initial reasoning prefixes (i.e., the first 32 tokens) sampled per question and fine-tune the model using LoRA adapters applied to the query and value matrices.  We then evaluate the downstream effects on test-time scaling performance across both low-resource languages and high-resource counterparts. We show the results in Table \ref{tab:ft_resource_qwen}. The training logs and the exact LoRA configuration can be found in Appendix \ref{sec:fine_tuning_details}.
\par

\section{Results}

\subsection{Language-Specific Test-Time Scaling in Strictly Monolingual Context}

In strict monolingual contexts, both \texttt{DeepSeek-R1-Distill-Llama-8B} and \texttt{DeepSeek-R1-Disill-Qwen-7B} exhibit a positive correlation between the number of generated tokens and reasoning accuracy, especially for the set of high-resource languages. Figure \ref{fig:overall_resource} shows that the test-time scaling trend is more prominent in \texttt{DeepSeek-R1-Distill-Llama-8B} as compared to \texttt{DeepSeek-R1-Distill-Qwen-7B}, while the latter exhibits a significantly inconsistent test-time scaling trend, especially for the set of low-resource languages. Figure \ref{fig:overall_resource} also highlights this disparity in the test-time scaling behavior through the lens of language resource availability. We observe that high-resource languages register a more significant gain in test-time scaling, while low-resource languages have less correlation of accuracy versus the amount of test-time inference compute.

\subsection{Language Fidelity in Reasoning Chains}
Deep diving into the quality of reasoning stream corresponding to each language, we test for any leakages into a foreign language. Figure \ref{fig:lang_id_2} shows the results for the two models under study. We observe that the `thought-language' appears to change to `English' in strictly non-English monolingual contexts. This phenomenon is highly prevalent in \texttt{DeepSeek-R1-Distill-Llama-8B} as compared to \texttt{DeepSeek-R1-Distill-Qwen-7B}, which maintains the fidelity of the target language. Moreover, this phenomenon is seen to occur more in low-resource languages as compared to high-resource languages, indicating the difficulty in maintaining long reasoning chains in a low-resource language.

\subsection{Similarity Analysis of Initial Reasoning Segments}
We conduct a series of exploratory experiments to quantify differences in initial thought patterns across languages.
% Figure \ref{fig:similarity_initial_reasoning} illustrates the similarity of the initial reasoning segments pertaining to different languages to that of English for \texttt{DeepSeek-R1-Distill-Llama-8B} on \texttt{AIME-Contest-2025-I}A. 
Figure \ref{fig:similarity_initial_reasoning}(a) shows results using the multilingual embedder \texttt{paraphrase-multilingual-mpnet-base-v2} while Figure \ref{fig:similarity_initial_reasoning}(b) shows results after translating all reasoning segments to English and using \texttt{all-mpnet-base-v2} to measure similarity. Low-resource languages have increasingly dissimilar initial thoughts compared to English and other high-resource languages. This suggests a structural divergence in early reasoning strategies depending on language resource availability. The two-pronged evaluation methodology confirms that these differences are not the result of embedding space mismatch or translation quality, but instead reflect deeper representational misalignment in multilingual model reasoning.

\subsection{Intra- and Inter-Language Consistency of Reasoning Prefixes}

To further understand how the base steps of creating a reasoning solution differ for different languages, we sample 100 extremely initial reasoning response prefixes (first 32 tokens) for each language and question pair. We measure the similarity of these reasoning solution prefixes by translating non-English prefixes to English and then utilizing \texttt{all-mpnet-base-v2}. The results in Figure \ref{fig:intra_prefix_similarity} reveal that low-resource languages tend to have lower intra-language similarity, indicating more variable early reasoning patterns, especially in \texttt{DeepSeek-R1-Distill-Qwen-7B}. In contrast, high-resource languages exhibit tighter and more consistent initial reasoning behaviors. These findings suggest that models demonstrate more stable inductive priors for high-resource languages.\par
For any two language pairs, we also measure the similarity between these reasoning prefixes and showcase the result averaged across all questions in Figure \ref{fig:inter_prefix_similarity} (Appendix). The heatmaps reveal a clear pattern: high-resource languages exhibit stronger mutual similarity, indicating convergent early reasoning structures. In contrast, low-resource languages show weaker alignment, both with each other and with high-resource languages, especially in \texttt{DeepSeek-R1-Distill-Qwen-7B}.

\subsection{Multilingual Initial Thought Transfer}

Motivated by previous trends of prefix similarity between high-resource and low-resource languages, we test if unsupervised fine-tuning on high-resource reasoning prefixes translates to enhancement of test-time scaling behavior. Figure \ref{fig:resource_qwen_e3} shows the results after training on each of 100 English reasoning prefixes for each question for both datasets. Unlike the inconsistent or flat trends seen in the base \texttt{DeepSeek-R1-Distill-Qwen-7B}, the fine-tuned model exhibits a clear upward trajectory with reasoning length — a hallmark of effective test-time scaling. Notably, this improvement is seen across both high-resource and low-resource languages, suggesting that English prefix tuning provides broadly transferable inductive priors for reasoning across the multilingual spectrum. We report similar performance gains after exposure to combined reasoning prefixes from all high-resource languages in Table \ref{tab:ft_resource_qwen}.

\begin{table}[h]
\centering
\begin{adjustbox}{max width=0.8\columnwidth}
\begin{tabular}{@{}lcrrrr}
\toprule
\textbf{Language} & \textbf{Strategy} & \multicolumn{4}{c}{\textbf{Maximum Token Limits}} \\
\cmidrule(lr){3-6}
 &  & \textbf{2000} & \textbf{4000} & \textbf{6000} & \textbf{8000} \\
\midrule
\multirow{3}{*}{English} 
 & Base & 0.164 & 0.123 & 0.175 & 0.178 \\
 & E-3  & \textbf{0.257} & \textbf{0.267} & \textbf{0.381} & \textbf{0.344} \\
 & H-1  & 0.133 & 0.083 & 0.189 & 0.275 \\
\addlinespace
\multirow{3}{*}{Portuguese}
 & Base & 0.186 & 0.210 & 0.205 & 0.141 \\
 & E-3  & 0.113 & 0.227 & \textbf{0.316} & \textbf{0.340} \\
 & H-1  & \textbf{0.181} & \textbf{0.254} & 0.251 & 0.328 \\
\addlinespace
\multirow{3}{*}{Italian}
 & Base & 0.096 & 0.152 & 0.146 & 0.171 \\
 & E-3  & 0.188 & 0.236 & 0.326 & \textbf{0.347} \\
 & H-1  & \textbf{0.210} & \textbf{0.332} & \textbf{0.343} & 0.271 \\
\addlinespace
\multirow{3}{*}{German}
 & Base & 0.138 & 0.156 & 0.222 & 0.191 \\
 & E-3  & 0.082 & \textbf{0.179} & \textbf{0.340} & \textbf{0.283} \\
 & H-1  & \textbf{0.250} & 0.084 & 0.000 & 0.000 \\
\addlinespace
\multirow{3}{*}{Vietnamese}
 & Base & 0.069 & 0.087 & 0.106 & 0.091 \\
 & E-3  & 0.133 & \textbf{0.239} & 0.235 & 0.162 \\
 & H-1  & \textbf{0.220} & 0.193 & \textbf{0.237} & \textbf{0.209} \\
\addlinespace
\multirow{3}{*}{Tagalog}
 & Base & 0.043 & 0.093 & 0.108 & 0.109 \\
 & E-3  & 0.077 & \textbf{0.171} & 0.176 & \textbf{0.349} \\
 & H-1  & \textbf{0.078} & 0.108 & \textbf{0.193} & 0.147 \\
\bottomrule
\end{tabular}
\end{adjustbox}
\caption{Accuracy scores across three strategies: Base (No Fine-tuning), E-3 (Fine-tuning on English prefixes for 3 epochs), and H-1 (Fine-tuning on high resource language prefixes: \{English, Portuguese, German, Italian\} for 1 epoch)—for six target languages at varying max token limits. \textbf{Bold} values indicate the best-performing strategy for each language at a given token length.}
\label{tab:ft_resource_qwen}
\end{table}

\vspace{-0.5cm}
\section{Discussion}

\texttt{DeepSeek-R1-Distill-Llama-8B} exhibits robust test-time scaling across languages, whereas \texttt{DeepSeek-R1-Distill-Qwen-7B} shows inconsistent or less significant test-time scaling, especially for low-resource languages. A probable cause is the difference in the pretraining composition of the two models: \texttt{Llama}'s natural multilingual corpus vs. \texttt{Qwen}'s code-heavy corpus and focused instruction following post-training. A more detailed examination of probable causes is left for a future study.\par
Despite being prompted and constrained to generate in a specific target language, models, especially \texttt{DeepSeek-R1-Distill-Llama-8B}, frequently switch mid-reasoning to English. This effect is especially pronounced in low-resource languages. When reasoning becomes complex or abstract, models revert to their most confident latent representations, which, due to pretraining bias, are often in English. This ``cross-lingual leakage'' reveals that language grounding is fragile in current multilingual models. It also raises important concerns about multilingual trustworthiness, particularly in educational or communicative settings. We discover that low-resource languages have less similar initial progression of thoughts to English as compared to other high-resource languages. Existing models likely align reasoning pathways more closely with high-resource languages, embedding linguistic bias into the thought process.

\par
Even within the same language and question, low-resource languages show high variance in their initial thought patterns. Inconsistency translates to unpredictable performance, which is harmful for real-world multilingual deployments. Generalization in multilingual settings requires internal coherence, a dimension often ignored in current multilingual reasoning benchmarks.\par
Motivated by the differences in the initial reasoning thoughts between languages and the overall higher performance of high-resource languages, we propose a data-efficient and unsupervised framework, MITT (Multilingual Initial Thought Transfer) where we utilize reasoning prefixes in high-resource languages. Fine-tuning \texttt{DeepSeek-R1-Distill-Qwen-7B} via MITT improves performance and scaling gains across all languages. Prefix-based adaptation is a lightweight, language-agnostic tool for improving low-resource performance not only from a multilingual reasoning benchmark lens but also test-time scaling behavior.
% This suggests a new paradigm to boost multilingual test-time scaling and reasoning performance: train on just the initial stable reasoning structures and transfer across languages.
\par 
\vspace{-0.2cm}
\section{Conclusion}

This work investigates the limitations of test-time scaling in multilingual language models, revealing that its benefits are neither uniform across languages nor consistent across model architectures. Our key contributions lie in establishing that multilingual test-time scaling is deeply uneven, identifying reasoning-level divergences that are not captured by traditional accuracy metrics, and proposing a simple yet effective strategy to bridge the gap for low-resource languages. These findings challenge the assumption that multilingual language models inherently generalize reasoning across linguistic boundaries. We argue that future multilingual models must prioritize reasoning fidelity and not just fluency. As test-time scaling becomes central to prompting, making it accessible across languages is both a technical priority and a matter of fairness. This work invites the community to look deeper at not just what models say, but what they think.
% --> Hiding conclusion for the time being!
\newpage
\section{Limitation}

While this work presents the first in-depth analysis of multilingual test-time scaling and proposes MITT as a simple and effective intervention, it comes with a few limitations. Our experiments are limited to Latin-script languages, and we do not test how the findings transfer to non-Latin scripts like Arabic, Hindi, or Chinese. We also evaluate only two decoder-only models, so we do not yet know how these results hold for other architectures like encoder-decoders or mixtures-of-experts. Although MITT uses prefixes from multiple high-resource languages, we have not tested whether reasoning patterns from low-resource languages can also be transferred or reused. Finally, our evaluation focuses on accuracy and reasoning similarity, but does not capture deeper aspects like factuality or logical soundness within the full reasoning chain.
\bibliography{custom}
\clearpage
\newpage
\appendix
\section{Multilingual Initial Thought Transfer (MITT): Fine-tuning details}\label{sec:fine_tuning_details}

To efficiently fine-tune \texttt{DeepSeek-R1-Distill-Qwen-7B}, we adopt a parameter-efficient training setup by combining 4-bit quantization with LoRA-based adapter tuning. We first load the model in 4-bit precision using the \texttt{BitsAndBytesConfig} from the \texttt{transformers} library, with computation performed in \texttt{bfloat16}. This enables memory-efficient fine-tuning while preserving performance.

We then apply LoRA adapters to the query and value projection layers (\texttt{"q\_proj"} and \texttt{"v\_proj"}) of the transformer blocks—these layers are known to be effective insertion points for instruction tuning in models. Our LoRA configuration uses a rank of $r = 8$, a scaling factor (\texttt{lora\_alpha}) of 32, and a dropout rate of 0.05. The bias is set to \texttt{"none"}, and the task type is specified as \texttt{"CAUSAL\_LM"} for autoregressive generation. This setup allows us to perform targeted adaptation while modifying only a small number of trainable parameters, making the approach both compute- and resource-efficient.

\begin{figure}[h]
    \centering
    \includegraphics[width=0.9\linewidth]{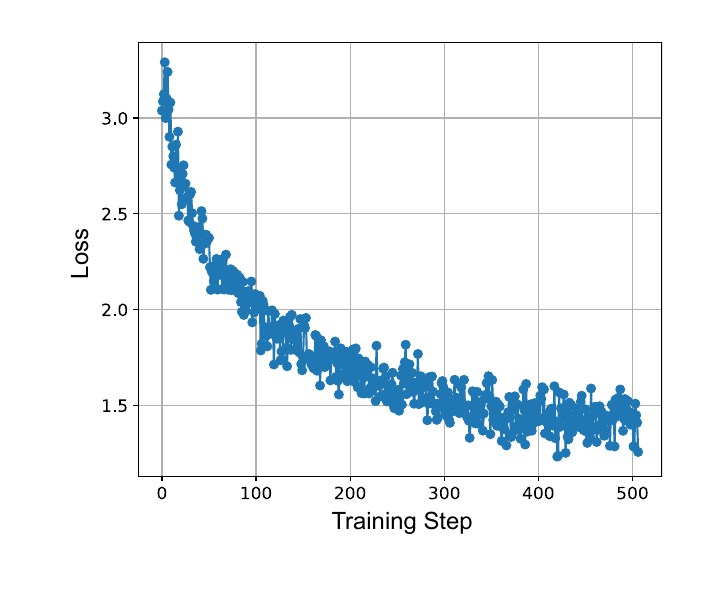}
    \vspace{-0.5cm}
    \caption{Training loss per training step for training on English reasoning prefixes for 3 epochs.}
    \label{fig:qwen_e3_loss}
\end{figure}
\vspace{-0.3cm}
\section{Prompt Details}\label{sec:prompt_details}

\begin{figure*}
    \centering
    \includegraphics[width=\linewidth]{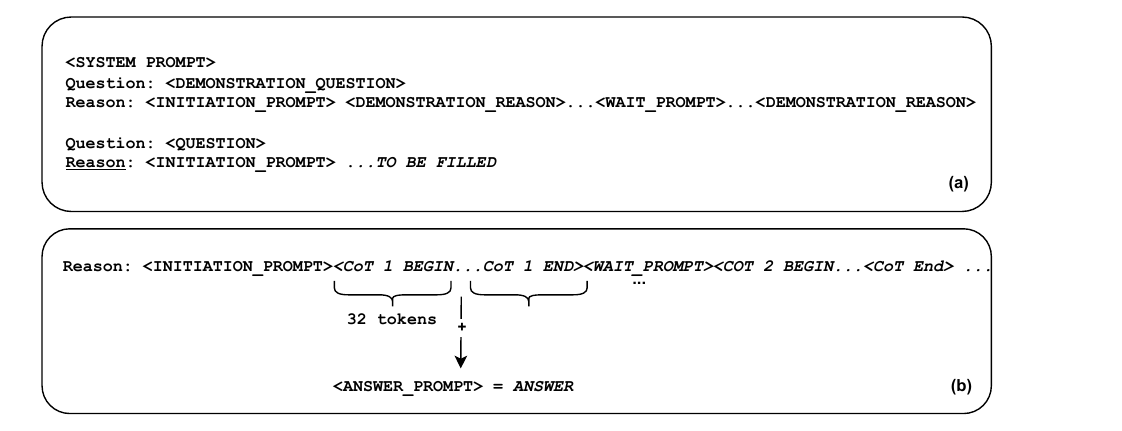}
    \caption{Panel (a) provides the full prompt template that we use for inference. Panel (b) gives an overview of our practice of placing the \texttt{WAIT\_PROMPT}s and \texttt{FINAL\_PROMPT}s}.
    \label{fig:prompt}
\end{figure*}

Figure \ref{fig:prompt} shows the prompt template we use for inferring at varied token lengths. We also show the mechanism of retrieving answers from the current reasoning chain using the \texttt{ANSWER\_PROMPT}s. We enforce the model to extend/re-think its reasoning chain of thought by manually inserting language-specific \texttt{WAIT\_PROMPT}s. The full list of language-specific \texttt{WAIT\_PROMPT}s, \texttt{ANSWER\_PROMPT}s and \texttt{SYSTEM\_PROMPT}s is provided in the Tables \ref{tab:wait_prompt},\ref{tab:answer_prompt} and  \ref{tab:system_prompt}.\par
Moreover, we use a set of \texttt{INITIATION\_PROMPT}s like `Let us think step-by-step in English' in English to impose the model to provide step-by-step reasoning. A full list of these \texttt{INITIATION\_PROMPT}s is provided in the table \ref{tab:step_tokens}.\par We use a 1-shot setting where we curate the reasoning for the question: $$\text{\textit{What is the value of the Gaussian integral}}$$ $$ \int_{-\infty}^{\infty} e^{-x^2} \, dx?$$ using \texttt{GPT-4o} followed by manually checking for the reasoning correctness and coherence. Following the completion of one CoT from the model, we manually put in the wait prompt and force the model to re-validate its thinking and produce one more end-to-end CoT. By curating the demonstration in this manner, we force the model to rethink its approach following the encounter of wait prompts. For different language scenarios, we present the same demonstration but in different languages, created by translation using \texttt{GPT-4o} as the zero-shot translator.
\vspace{-0.3cm}
\begin{table*}[h]
\small
\centering
\begin{adjustbox}{max width=\textwidth}
\begin{tabular}{@{}p{3cm}p{12cm}@{}}
\toprule
\textbf{Language} & \textbf{Wait Prompt} \\
\midrule
\textbf{English}     & Let me re-think my reasoning from scratch. \\
\addlinespace
\textbf{German}      & Lass mich mein Denken von Grund auf neu überdenken. \\
\addlinespace
\textbf{Portuguese}  & Deixe-me repensar meu raciocínio do zero. \\
\addlinespace
\textbf{Italian}     & Fammi ripensare il mio ragionamento da capo. \\
\addlinespace
\textbf{Vietnamese}  & {Hãy để tôi suy nghĩ lại từ đầu.} \\
\addlinespace
\textbf{Tagalog}     & {Hayaan mo akong muling pag-isipan ang aking pangangatwiran mula sa simula.} \\
\bottomrule
\end{tabular}
\end{adjustbox}
\centering
\caption{Wait prompts used for extending chain of thoughts, in different languages.}
\label{tab:wait_prompt}
\end{table*}

\begin{table*}[h]
\small
\centering
\begin{adjustbox}{max width=\textwidth}
\begin{tabular}{@{}p{3cm}p{12cm}@{}}
\toprule
\textbf{Language} & \textbf{Initiation Prompt} \\
\midrule
\textbf{English}     & Let us think step-by-step in English. \\
\addlinespace
\textbf{German}      & Lass uns Schritt für Schritt auf Deutsch denken. \\
\addlinespace
\textbf{Italian}     & Pensiamo passo dopo passo in italiano. \\
\addlinespace
\textbf{Portuguese}  & Vamos pensar passo a passo em português. \\
\addlinespace
\textbf{Vietnamese}  & {Hãy suy nghĩ từng bước bằng tiếng Việt.} \\
\addlinespace
\textbf{Tagalog}     & {Mag-isip tayo nang sunud-sunod sa Tagalog.} \\
\bottomrule
\end{tabular}
\end{adjustbox}
\centering
\caption{Initiation prompts used for initiating the chain of thoughts in different languages.}
\label{tab:step_tokens}
\end{table*}

\begin{table*}[h]
\small
\centering
\begin{adjustbox}{max width=\textwidth}
\begin{tabular}{@{}p{3cm}p{12cm}@{}}
\toprule
\textbf{Language} & \textbf{Answer Prompt} \\
\midrule
\textbf{English} & 
Stop thinking. What is the final answer? Output only the raw number in the format \verb|\( \boxed{} \)|. No text, no symbols, no punctuation. Just the number.

Answer: \\
\addlinespace
\textbf{German} & 
Hör auf zu denken. Was ist die endgültige Antwort? Gib nur die reine Zahl im Format \verb|\( \boxed{} \)| aus. Kein Text, keine Symbole, keine Satzzeichen. Nur die Zahl.

Antwort: \\
\addlinespace
\textbf{Italian} & 
Smetti di pensare. Qual è la risposta finale? Mostra solo il numero grezzo nel formato \verb|\( \boxed{} \)|. Nessun testo, nessun simbolo, nessuna punteggiatura. Solo il numero.

Risposta: \\
\addlinespace
\textbf{Portuguese} & 
Pare de pensar. Qual é a resposta final? Mostre apenas o número bruto no formato \verb|\( \boxed{} \)|. Sem texto, sem símbolos, sem pontuação. Apenas o número.

Resposta: \\
\addlinespace
\textbf{Vietnamese} & 
{Dừng suy nghĩ. Đáp án cuối cùng là gì? Chỉ xuất ra con số thô ở định dạng \verb|\( \boxed{} \)|. Không có văn bản, ký hiệu hoặc dấu câu. Chỉ số thôi.

Đáp án:} \\
\addlinespace
\textbf{Tagalog} & 
{Itigil ang pag-iisip. Ano ang huling sagot? I-output lamang ang hilaw na numero sa format na \verb|\( \boxed{} \)|. Walang teksto, walang simbolo, walang bantas. Numero lamang.

Sagot:} \\
\bottomrule
\end{tabular}
\end{adjustbox}
\caption{Answer prompts used to extract answer from current state of reasoning, in different languages.}
\label{tab:answer_prompt}
\end{table*}

\begin{table*}[h]
\small
\centering
\begin{adjustbox}{max width=\linewidth}
\begin{tabular}{@{}p{2.5cm}p{13cm}@{}}
\toprule
\textbf{Language} & \textbf{System Prompt} \\
\midrule
\textbf{English} & You are a mathematical question-answering assistant. Always use English for all reasoning, explanations, and answers. Never switch to any other language at any point. A mathematical question will be provided in English. First, give a detailed explanation in English that leads to the solution. Then, output the final answer using the format \verb|\( \boxed{...} \)| — but only after the full reasoning is completed in English. Do not translate anything. Maintain full linguistic consistency in English from start to finish. This rule must never be broken. \\
\addlinespace
\textbf{German} & Du bist ein Assistent zur Lösung mathematischer Aufgaben. Verwende immer Deutsch für alle Überlegungen, Erklärungen und Antworten. Wechsle niemals zur englischen Sprache oder einer anderen Sprache. Dir wird eine mathematische Frage auf Deutsch gestellt. Erkläre zunächst ausführlich auf Deutsch, wie man zur Lösung kommt. Dann gib die endgültige Antwort im Format \verb|\( \boxed{...} \)| an — aber nur, nachdem die gesamte Begründung auf Deutsch abgeschlossen ist. Übersetze nichts ins Englische. Bleibe vollständig bei der deutschen Sprache. Diese Regel darf niemals gebrochen werden. \\
\addlinespace
\textbf{Italian} & Sei un assistente per la risoluzione di problemi matematici. Usa sempre l'italiano per tutti i ragionamenti, le spiegazioni e le risposte. Non passare mai all'inglese o a un'altra lingua. Ti verrà fornita una domanda matematica in italiano. Fornisci prima una spiegazione dettagliata in italiano che conduca alla soluzione. Poi, fornisci la risposta finale usando il formato \verb|\( \boxed{...} \)|, ma solo dopo che il ragionamento completo è stato fornito in italiano. Non tradurre nulla in inglese. Mantieni la coerenza linguistica completa in italiano dall'inizio alla fine. Questa regola non deve mai essere infranta. \\
\addlinespace
\textbf{Portuguese} & Você é um assistente de resolução de questões matemáticas. Sempre use o português para todos os raciocínios, explicações e respostas. Nunca mude para o inglês ou qualquer outro idioma. Uma pergunta matemática será fornecida em português. Primeiro, forneça um raciocínio detalhado em português que leve à solução. Depois, apresente a resposta final usando o formato \verb|\( \boxed{...} \)|, mas somente após concluir todo o raciocínio em português. Não traduza nada para o inglês. Mantenha total consistência linguística em português do início ao fim. Esta regra nunca deve ser quebrada. \\
\addlinespace
\textbf{Vietnamese} & {Bạn là một trợ lý giải toán. Luôn sử dụng tiếng Việt cho toàn bộ phần lập luận, giải thích và câu trả lời. Tuyệt đối không chuyển sang tiếng Anh hoặc bất kỳ ngôn ngữ nào khác. Một câu hỏi toán học sẽ được đưa ra bằng tiếng Việt. Trước tiên, hãy giải thích chi tiết bằng tiếng Việt dẫn đến cách giải. Sau đó, đưa ra đáp án cuối cùng bằng định dạng \verb|\( \boxed{...} \)| — nhưng chỉ sau khi phần giải thích hoàn toàn bằng tiếng Việt đã được trình bày. Tuyệt đối không dịch bất cứ phần nào sang tiếng Anh. Phải giữ sự nhất quán ngôn ngữ tiếng Việt từ đầu đến cuối. Đây là nguyên tắc bắt buộc.} \\
\addlinespace
\textbf{Tagalog} & {Ikaw ay isang assistant na sumasagot sa mga tanong sa matematika. Palaging gumamit ng Tagalog para sa lahat ng pangangatwiran, paliwanag, at sagot. Huwag kailanman lumipat sa anumang ibang wika sa kahit anong punto. Ang isang tanong sa matematika ay ibibigay sa Tagalog. Una, magbigay ng detalyadong paliwanag sa Tagalog na humahantong sa solusyon. Pagkatapos, ibigay ang panghuling sagot gamit ang format na \verb|\( \boxed{...} \)| — ngunit gawin ito lamang pagkatapos makumpleto ang buong paliwanag sa Tagalog. Huwag magsalin ng anuman. Panatilihin ang buong pagkakapare-pareho ng wika sa Tagalog mula simula hanggang wakas. Ang panuntunang ito ay hindi dapat labagin kailanman.} \\
\bottomrule
\end{tabular}
\end{adjustbox}
\caption{System instruction prompts used for inference, in different languages.}
\label{tab:system_prompt}
\end{table*}

\section{Intra- and Inter-Language Consistency of Reasoning Prefixes}\label{sec:intra_and_inter}
\begin{figure*}[ht]
    \centering
    \includegraphics[width=\linewidth]{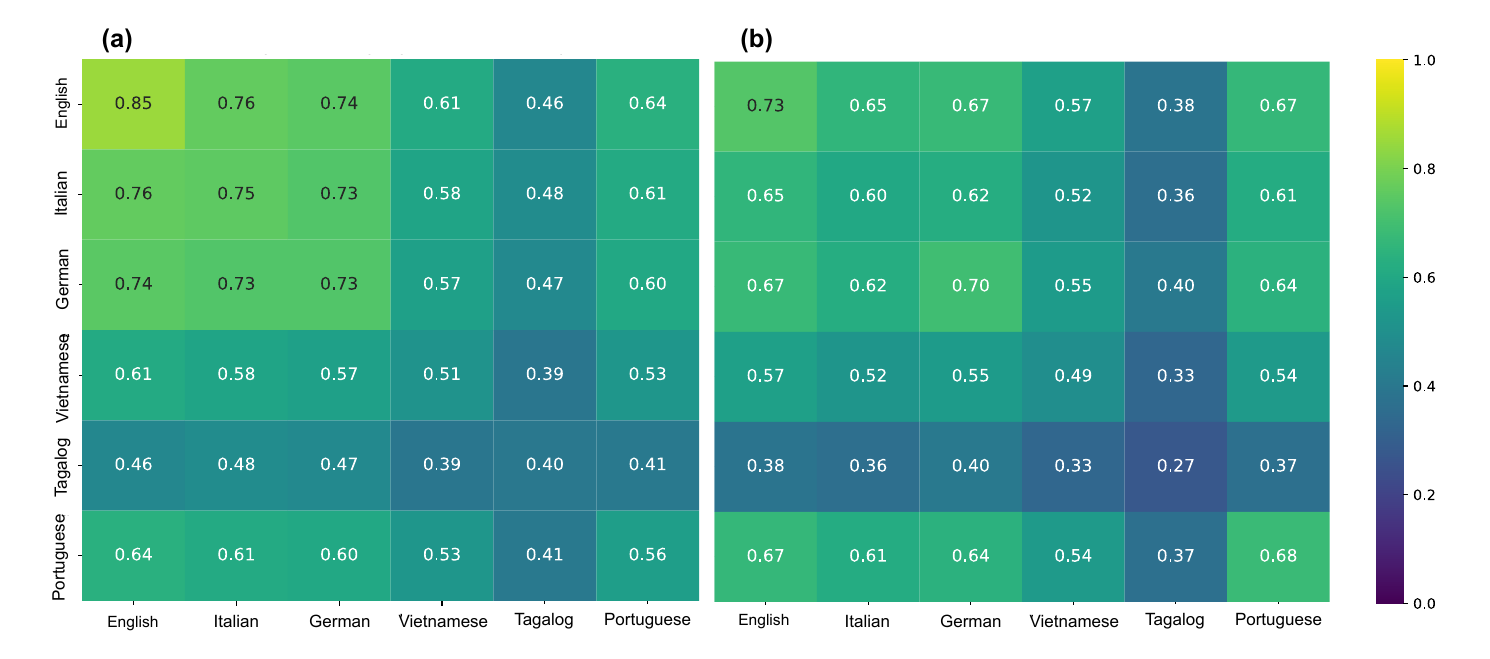}
    \caption{Figure shows pairwise similarity scores between initial reasoning segments (first 32 tokens) generated in different languages, averaged across questions. Each cell represents the mean similarity between two language-specific reasoning distributions. Panel (a) corresponds to \texttt{DeepSeek-R1-Distill-LLama-8B} and panel (b) to \texttt{DeepSeek-R1-Distill-Qwen-7B.} }
    \label{fig:inter_prefix_similarity}
\end{figure*}

To analyze cross-lingual reasoning behavior, we compute the average similarity of reasoning prefixes for all language pairs, as shown in Figure \ref{fig:inter_prefix_similarity}. The heatmaps uncover a notable trend: high-resource languages tend to share more similar early reasoning structures, reflecting a degree of convergence. In contrast, low-resource languages exhibit limited alignment—both with each other and with high-resource counterparts—most prominently in \texttt{DeepSeek-R1-Distill-Qwen-7B}.
\section{Language Fidelity in Reasoning Chains}\label{sec:language_fidelity}

\begin{figure*}[h]
    \centering
    \includegraphics[width=\linewidth]{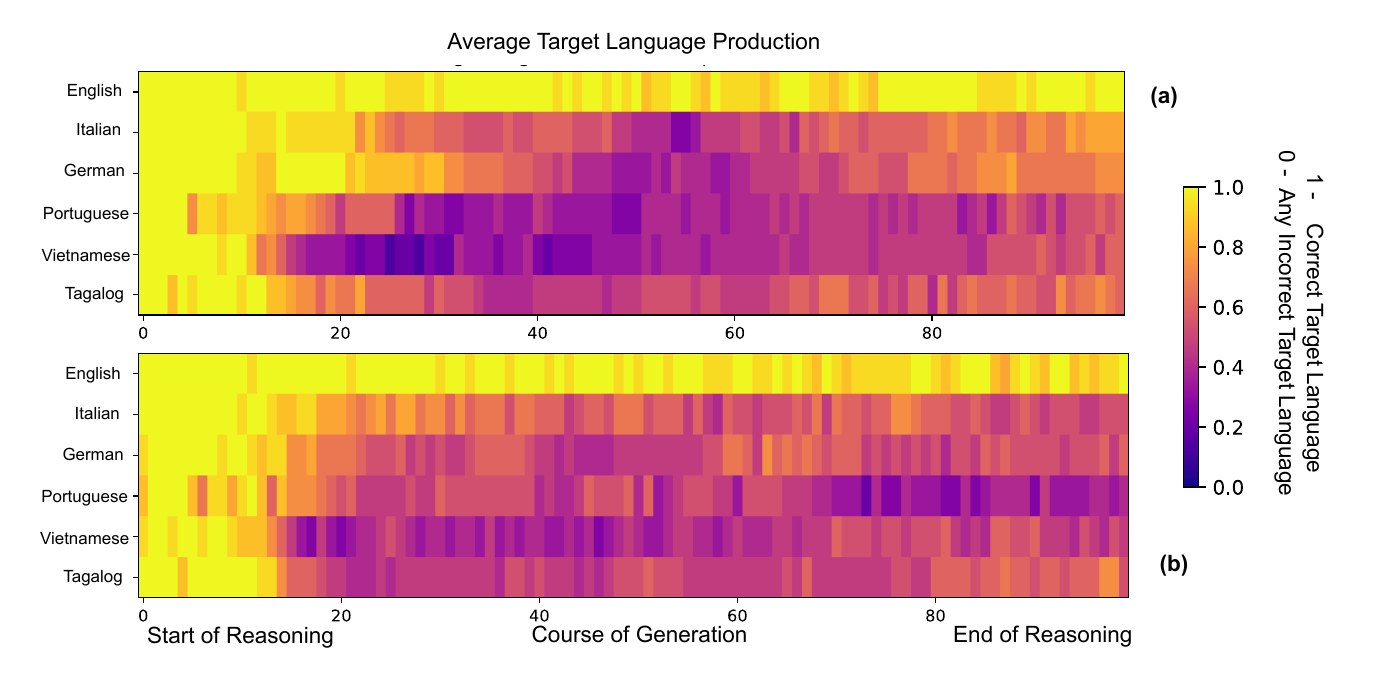}
    \caption{Each row represents the average success score in target language production per token generation segment. We create these segments in a similar manner as we create them for Figure \ref{fig:lang_id_2}. Panels \textbf{(a)} and \textbf{(b)} corresponding to \texttt{DeepSeek-R1-Distill-LLama-8B} and \texttt{DeepSeek-R1-Distill-Qwen-7B}, respectively. While the models are prompted monolingually in the target language, we observe significant cross-lingual leakage.}
    \label{fig:lang_id_2_granular}
\end{figure*}

We present a more fine-grained analysis of `cross-lingual leakage' by quantifying the average number of successes of token generation in the target language of our monolingual setting. In place of the 20-bin downsampled representation shown in Figure \ref{fig:lang_id_2}, we construct a more fine-grained 100-bin version for improved resolution. We show the heatmap for \texttt{DeepSeek-R1-Llama-8B} and \texttt{DeepSeek-R1-Qwen-7B} in Figure \ref{fig:lang_id_2_granular}(a) and Figure \ref{fig:lang_id_2_granular}(b). We observe that apart from English, there is an increased production of non-target language production as the target language changes from high-resource to low-resource.

\end{document}